\documentclass[10pt,twocolumn,letterpaper]{article}

\newif\ifarxiv

\arxivtrue

\ifarxiv
    \usepackage[pagenumbers]{cvpr} 
\else
    \usepackage[review]{cvpr}      
\fi
\usepackage{booktabs}
\usepackage{tabularx}
\usepackage{amsfonts, amsmath, amssymb, amsthm}
\usepackage{stmaryrd}
\usepackage{ulem}
\usepackage{siunitx}

\newcommand{\mathbbm}[1]{\mathbf{#1}}

\newcommand{\reals}[0]{\mathbb{R}}

\newcommand{\bouleps}[0]{\mathcal{B}_{p}^{\epsilon}}

\newtheorem{definition}{Definition}

\newtheorem{proposition}{Proposition}

\definecolor{cvprblue}{rgb}{0.21,0.49,0.74}
\usepackage[pagebackref,breaklinks,colorlinks,allcolors=cvprblue]{hyperref}

\def\paperID{11617} 
\def\confName{CVPR}
\def\confYear{2026}

\title{Fast and Flexible Robustness Certificates for Semantic Segmentation}

\author{
    Thomas Massena\textsuperscript{1,3} \and
    Corentin Friedrich\textsuperscript{2} \and
    Franck Mamalet\textsuperscript{2} \and
    Mathieu Serrurier\textsuperscript{1}
}

\begin{document}

\maketitle

\ifarxiv
    {
        \renewcommand{\thefootnote}{}
        \footnotetext[1]{\textsuperscript{1}Institut de Recherche en Informatique de Toulouse}
        \footnotetext[2]{\textsuperscript{2}IRT Saint-Exupery}
        \footnotetext[3]{\textsuperscript{3}SNCF, DTIPG}
    }
\else 
\fi

\ifarxiv
    
\begin{abstract}
    Deep Neural Networks are vulnerable to small perturbations that can drastically alter their predictions for perceptually unchanged inputs. The literature on adversarially robust Deep Learning attempts to either enhance the robustness of neural networks (e.g, via adversarial training) or to certify their decisions up to a given robustness level (e.g, by using randomized smoothing, formal methods or Lipschitz bounds). These studies mostly focus on classification tasks and few efficient certification procedures currently exist for semantic segmentation. In this work, we introduce a new class of certifiably robust Semantic Segmentation networks with built-in Lipschitz constraints that are efficiently trainable and achieve competitive pixel accuracy on challenging datasets such as Cityscapes. Additionally, we provide a novel framework that generalizes robustness certificates for semantic segmentation tasks, where we showcase the flexibility and computational efficiency of using Lipschitz networks. Our approach unlocks real-time compatible certifiably robust semantic segmentation for the first time. Moreover, it allows the computation of worst-case performance under $\ell_2$ attacks of radius $\epsilon$ across a wide range of performance measures. Crucially, we benchmark the runtime of our certification process and find our approach to be around 600 times faster than randomized smoothing methods at inference with comparable certificates on an NVIDIA A100 GPU. Finally, we evaluate the tightness of our worst-case certificates against state-of-the-art adversarial attacks to further validate the performance of our method.
\end{abstract}

    \section{Introduction}

Deep Neural Networks (DNNs) have transformed the landscape of machine learning, driving breakthroughs across perception, reasoning, and decision-making tasks. 
Yet, their remarkable expressivity comes at the cost of a vulnerability to carefully crafted adversarial perturbations~\cite{christian2014intriguing}. These weaknesses raise critical concerns when deploying Machine Learning (ML) systems in safety-critical contexts such as autonomous driving or healthcare, where guarantees on the reliability of decisions are crucial \cite{yin2025certified}.

While obtaining robustness certificates for tasks other than classification has been attempted in the context of formal verification methods~\cite{cohen2024verification} and randomized smoothing~\cite{fischer2021scalable, chiang2020detection}. Lipschitz-by-design methods, introduced in~\citet{tsuzuku2018lipschitz}, were never adapted to provide downstream certificates for more complex tasks than classification or regression. Our contributions are as follows:

\begin{itemize}
    \item First, we introduce a general class of robustness certificates for segmentation tasks, tailored to certify the robustness of a variety of performance measures to general adversarial degradation objectives. We propose certificates for worst case performance under attack and minimum adversarial budgets required to satisfy an adversarial objective.
    \item Then, using this framework, we build certificates for the Pixel Accuracy, FNR or Class IoU performance measures under attack using the Lipschitz constant of a model.
    \item Additionally, we develop the first efficiently trainable Lipschitz-constrained neural network models for safety-critical semantic segmentation tasks using a DeepLabV3-like architecture, suitable for challenging tasks like the Cityscapes dataset.
    \item Leveraging our framework along with Lipschitz networks,
    we compute robustness certificates for several safety-critical scenarios. Most importantly, our networks are suitable for real-time inference and we provide the first certifiably robust method that allows the computation of robustness certificates in less than $0.1$ seconds on $1024$ by $1024$ images from the Cityscapes dataset. 
\end{itemize}

\noindent Additionally we provide a code-base available on \href{https://github.com/deel-ai-papers/robust-segmentation}{GitHub}, that allows the training and certification of Lipschitz-constrained neural networks in $\ell_2$ norm on segmentation tasks.

    \section{Background on Adversarial Robustness}

In this section, we will introduce some key concepts and their formulations in the context of classification tasks. Then, we will present relevant certifiable robustness works in the context of semantic segmentation. 

\paragraph{Notations} In our settings, we will denote $\mathcal{X}$ as the input space, $\mathcal{Y}$ as the output space, $\mathcal{K}$ as the label space and $\Omega$ as the coordinate space (e.g the images, masks, classes and pixel indexes respectively). 
In the following sections, we denote a general semantic segmentation model as $f : \mathcal{X} \rightarrow \mathcal{Y}, \ \text{with} \ \mathcal{X} := [0, 1]^{|\Omega|} ,\ \mathcal{Y} := \reals^{|\mathcal{K}| \times |\Omega|}$ and $|\cdot|$ the cardinality measure of a set. More specifically, in certain cases, we will refer to dimension sizes $K, \ H$ and $W$ which stand for the class, height and width dimensions of images respectively. Also, we denote $\bouleps(x)$ as the $p$-norm ball of radius $\epsilon$ around $x$.
Finally, we denote the output of a segmentation model as $f(X)$, its coordinate-wise predictions (e.g for pixels) as $f(X)_\omega \in \reals^{|\mathcal{K}|}$ and $f^k(X)_\omega$ as the output associated to class $k \in \mathcal{K}$ for coordinate $\omega \in \Omega$. Finally, we denote the predicted class by a semantic segmentation network at coordinate $\omega \in \Omega$ as $\hat{Y}_\omega = \arg \max_{k \in \mathcal{K}} f^k(X)_\omega$, which becomes $\tilde{Y}_\omega = \arg \max_{k \in \mathcal{K}} f^k(\tilde{X})_\omega$ for an attacked input $\tilde{X} \in \bouleps(X)$.

\subsection{Adversarial Robustness in the Classification Setting}

Seminal adversarial deep learning works originally mostly focused on designing adversarial attacks to fool deep neural classification models~\cite{christian2014intriguing}. To this day, a sizeable proportion of adversarial deep learning works still focus on designing maximally efficient attacks on classification tasks. In the following paragraphs, we define the key concepts surrounding certifiable adversarial robustness in the context of classification tasks.

\begin{definition}[Robustness radius - classification]
    For any predictive model, $f : \mathcal{X} \rightarrow \reals^{|\mathcal{K}|}$ and any test point $(X,Y)$, we define:

    \begin{equation}
    \begin{aligned}
        R (X,Y) = \inf &\{\epsilon \in \reals^+ | 
        \exists \tilde{X} \in \bouleps(X),  \\
        & \arg \max_{k \in \mathcal{K}} f^k(\tilde{X}) \neq Y\}.
    \end{aligned}
    \end{equation}

    \noindent as the robustness radius of the model prediction. 
\label{def:classif_radius}
\end{definition}

This definition is widely adopted in the context of certifiably robust classification and allows to characterize how ``invariant'' a neural network's prediction is to local perturbations. In practice, there are multiple ways of obtaining a lower bound on the value of the robustness radius as defined in Definition~\ref{def:classif_radius}.
We can distinguish three main competing approaches to lower-bound the robustness radius of a DNN's prediction locally. Namely, formal-verification methods, randomized smoothing based variants, and Lipschitz neural networks stand out as the most popular approaches.

\begin{itemize}
    \item \textbf{Formal verification methods:} allow for the computation of worst-case logit variations under $\epsilon$-bounded $\ell_p$-norm attacks~\cite{tran2021robustness} at inference by using formal-based solvers.
    \item \textbf{Randomized smoothing methods:} which define a smoothed classifier $g$ that is estimated via Monte-Carlo methods~\cite{cohen2019certified,lecuyer2019certified}. This smooth classifier is then certifiable probabilistically when using well-chosen sampling distributions~\cite{yang20wulff}.
    \item \textbf{Lipschitz neural networks:} which leverage a by-design Lipschitz upper bound to compute worst-case logit variations with no inference time overhead.
\end{itemize}

Importantly, all of these methods allow the user to find a conservative lower bound $\underline{R}(X,Y) \leq R(X,Y)$ such that no adversarial attack inferior to $\epsilon$ in $\ell_p$ norm could misclassify the originally correctly classified sample $X$. These robustness lower bounds are usually coined as robustness \textit{certificates}.

\subsection{Lipschitz Certificates for Classification}

A function $f: \reals^m \rightarrow \reals^n$ is said to be $L$-Lipschitz in $\ell_p$ norm if it verifies:

\begin{equation}
    \forall (X_1, X_2) \in \reals^m, \| f(X_1) - f(X_2) \|_p \leq L . \| X_1 - X_2 \|_p
\label{def:lipschitz}
\end{equation}

with $L > 0$. This property effectively bounds how ``stable'' a predictor's decision remains under perturbations. Moreover, Lipschitz continuous neural networks are universal approximators like their unconstrained counterparts on tasks other than regression  (c.f Section 6 of \citet{anil2019sorting}).

\begin{proposition}[Classification robustness bound]
    For any $L$-Lipschitz predictive model in $\ell_p$ norm, $f : \reals^d \rightarrow \reals^{|\mathcal{K}|}$, we have the following lower bound on $R(X, Y)$:

    \begin{equation}
        R(X, Y) \geq \mathbbm{1}_{\hat{Y} = Y}. 2^\frac{1-p}{p} . \mathcal{M}_X(f) / L.
    \end{equation}

    \noindent with $\mathcal{M}_X(f) = f^\mathrm{top1}(X) - f^\mathrm{top2}(X)$, as explicited in \cite{li2019preventing}. Here, the $\mathbbm{1}_{\{\hat{Y} = Y \}}$ term ensures that the robustness radius of originally misclassified samples is zero.
\label{prop:classification-certificate}
\end{proposition}

We have that Lipschitz-constrained networks allow for easily computable robustness radius lower bound computations by using the global Lipschitz constant of a specially designed neural network.

\subsection{Related Works on Robust Segmentation}
\label{sec:related_works}

While adversarial robustness has been extensively investigated for image classification, the robustness of semantic segmentation models has received comparatively less attention. 
Early work by \citet{Arnab_2018_CVPR} provided one of the first systematic analyses of adversarial examples on modern semantic segmentation architectures, showing that dense prediction exhibits vulnerability patterns that differ from standard classification.
Building on segmentation-specific threat models, \citet{rony2023proximal} introduced a proximal-splitting-based adversarial attack that directly optimizes dense pixel-wise perturbations and yields substantially stronger white-box baselines than attacks adapted from classification, providing a stronger benchmark for assessing segmentation robustness. 
These works underline that semantic segmentation robustness remains relatively underexplored compared to the classification case and motivate further study in safety-critical domains such as automated driving perception, for which broader surveys of certified robustness have recently been proposed \cite{yin2025certified}. 
In the context of certifiably robust Semantic Segmentation the overall approaches to certification remain the same as for classification. However, the high dimensionality of the neural segmentation network's outputs complicate the analysis. We can denote two main existing approaches to robust semantic segmentation tasks.

\paragraph{Formal verification methods} While some formal-verification methods allow for the verification of a segmentation network's predictions on toy datasets~\cite{tran2021robustness}, formal-based certifiable methods lack scalability or end up providing vacuous bounds which are not applicable to over-parametrized networks that segment high resolution images. Indeed, these methods usually have time complexities that grow quadratically with the number of neurons and layers of the model~\cite{katz2017reluplex}. Therefore, even the most prestigious neural network verification competitions only include verification tasks for simple semantic segmentation networks with less than a million parameters~\cite{muller2022third}.
Currently, specific approaches to robust semantic segmentation using formal-based methods include patch-based attack defenses~\cite{luo2023formalbenchmark} and approaches that can segment MNIST-like images with pixel-wise certificates~\cite{pal2023benchmarkseg}.

\paragraph{Randomized Smoothing} In the context of semantic segmentation, \citet{fischer2021scalable} first applied randomized smoothing to certify subsets of pixel-wise predictions up to a unique robustness radius $R$ with probability $1-\alpha$. Indeed, the authors find that applying randomized smoothing naively to the $H \times W$ classifications provided by the segmentation network requires too many samples to certify the validity of all pixel classifications with probability $1-\alpha$. Therefore, they approach certification as a multiple-testing problem where the \textsc{SegCertify} method 
controls the family-wise error rate at level $\alpha$ using a step-down procedure to ensure the validity of the segmentation of a whole image up to a user-chosen error probability $\alpha$.  
To improve empirical performance while preserving the global confidence level, pixels whose smoothed predictions do not achieve the required confidence are assigned to an abstention set.
Unfortunately, this method requires running multiple forward passes for the model, which is even more cumbersome than in the classification setting, given that the memory requirements of segmentation models are usually high. Moreover, the computation of statistical processes for each pixel in the image also add a significant runtime overhead. Therefore, these methods are currently unusable for systems where predictions need to be inferred in generally less than a second.

More recently, the \textsc{LocalizedLP} method was introduced in \citet{schuchardtlocalized}. This method improves on the \textsc{SegCertify} method by leveraging the local dependencies of semantic segmentation networks via isotropic Gaussian smoothing. 
Unfortunately, this improvement comes at a price in terms of sample complexity as \textsc{LocalizedLP} uses around 15$\times$ more MC iterations that \textsc{SegCertify}, as explicited in Section 7.1 of \cite{schuchardtlocalized}. 
For reference, the implementation of \citet{schuchardtlocalized} uses a DeepLabV3 architecture similar to ours on the Cityscapes dataset, and reports 1204 seconds of runtime per image by using $153600$ Monte Carlo iterations. 
Therefore, we do not compare the efficiency of our method to the \textsc{LocalizedLP} method given that it is clearly not suited for fast inference scenarios.

    \section{Unifying Semantic Segmentation Robustness Metrics}
\label{sec:framework}

The notion of adversarial robustness in semantic segmentation is not clearly defined, and relies on sparse metrics introduced throughout the literature. Adversarial attacks usually aim to increase a proxy loss function with a fixed perturbation budget $\epsilon$, in order to decrease the pixel accuracy and the mIoU. In contrast, the attack introduced in~\cite{rony2023proximal} seeks the minimal budget to successfully perturb at least $\gamma \%$ of pixels.
Regarding defenses, some works focus on certifying pixel-wise classification~\cite{tran2021robustness,fischer2021scalable}, whereas others ensure collective robustness certificates for the classification of pixel subgroups~\cite{schuchardtlocalized}.
In this section, we present a general framework that unifies the notion of robustness, especially for semantic segmentation. 

\paragraph{Two paradigms of robustness}
In a safety-critical setting, robustness at a given sample $(X,Y)$ can be addressed from two different perspectives. We express these perspectives as the following questions.

\begin{quote}
    \textbf{Q1} --- \textit{Given an adversarial budget $\epsilon$, what is the worst performance I could reach?} \\
    \textbf{Q2} --- \textit{If I want to degrade the performance metric to satisfy a degradation objective $\kappa$, what adversarial budget do I need?}
\end{quote}

Most attacks and certifiable defenses try to answer \textbf{Q1} by degrading the usual segmentation metrics (i.e. pixel accuracy or mIoU) with a fixed budget. Answering \textbf{Q2} is less common. For example, \citet{rony2023proximal} consider the degradation of the pixel accuracy metric with the objective of less than $\gamma \%$.
In the following, we present two novel definitions to answer both questions: we define the \textit{worst-case performance} for \textbf{Q1}, and the \textit{generalized robustness radius} for \textbf{Q2}.

\begin{definition}[Worst-case performance]
    For any predictive model $f : \mathcal{X} \rightarrow \mathcal{Y}$, and performance metric $h: \mathcal{Y} \times \mathcal{K}^{|\Omega|} \rightarrow \reals$, we define the $\epsilon$~worst-case performance measured on a data point $(X,Y)$ as:

    \begin{equation}
       h_\epsilon(X,Y) = \min_{\tilde{X} \in \bouleps(X)} h \left(f(\tilde{X}), Y \right).
    \end{equation}

    In our setting, we assume that $h$ is positively correlated with system performance, i.e ``higher is better''.
\label{def:wc_performance}
\end{definition}

For example with $h$ being the pixel accuracy measure, $h_\epsilon$ is the minimum reachable pixel accuracy on an image and mask pair $(X,Y)$ under fixed adversarial budget $\epsilon$.

\begin{definition}[Generalized robustness radius]
    For any predictive model $f: \mathcal{X} \rightarrow \mathcal{Y}$, performance metric $h: \mathcal{Y} \times \mathcal{K}^{|\Omega|} \rightarrow \reals$, and degradation objective $\kappa: \reals \rightarrow \{0, 1\}$, we define the generalized robustness radius on a data point $(X,Y)$ as:

    \begin{equation}
    \begin{aligned}
        R_\kappa (X,Y) = \inf &\{ \epsilon \in \reals^+ | \ \exists \tilde{X} \in \bouleps(X), \\
        &\kappa \left[ h(f(\tilde{X}), Y) \right] = 1 \}.
    \end{aligned}
    \end{equation}

    \noindent The degradation objective $\kappa$ on performance metric $h$ is either unsatisfied (0=failure) or satisfied (1=success).
\label{def:generalized-robustness}
\end{definition}

This definition allows us to define a robustness radius for any desired adversarial degradation objective $\kappa$ on performance measure $h$ in a variety of different scenarios\footnote{Note that $\kappa$ could also be defined as $\kappa : \reals \times \reals \rightarrow \{ 0, 1\}$ with $\kappa[h(f(\tilde{X}),Y), h(f(X),Y)]$ to ensure robustness to degradations that are relative to the performance of the network on a clean point (e.g the accuracy must be at least half of the original accuracy).}. For example, with $h$ being the pixel accuracy, and $\kappa(z) = \mathbbm{1}_{\{z \leq\gamma\%\}}$ the degradation objective, the radius $R_\kappa$ is the minimum budget to reach a pixel accuracy below $\gamma\%$ under attack.

Computing exactly $h_\epsilon$ and $R_\kappa$ is infeasible in practice. In certified robustness, we answer \textbf{Q1} and \textbf{Q2} by computing lower bounds (also called certificates) on $h_\epsilon$ and $R_\kappa$. Section~\ref{sec:example_pacc} will illustrate how to compute those certificates for the pixel accuracy metric.
    \section{Fast Estimation of Robustness Certificates}

In the following Section, we show how to use Lipschitz certificates to give lower-bound approximates for questions \textbf{Q1} and \textbf{Q2} efficiently.

\paragraph{Computing certificates} In order to compute the worst-case performance variations of an $L$-Lipschitz predictor under $\epsilon$-bounded attacks, we leverage:
 
\begin{equation}
\begin{aligned}
     h_\epsilon (X) &= \min_{\delta \in \mathcal{B}_p^{\epsilon}(0)} h(f(X + \delta), Y) \\
     &\geq \min_{\alpha \in \mathcal{B}_p^{L\epsilon}(0)} h (f(X) + \alpha,Y).
\label{eq:min_perf}
\end{aligned}
\end{equation}

\noindent This conversion from input space perturbations in $\mathcal{B}_p^\epsilon(X)$ to output perturbations in $\mathcal{B}_p^{L\epsilon}(f(X))$ comes directly from the $L$-Lipschitz property of the neural network $f$ 
(for the sake of completeness a proof is given in Appendix A).
Note that this lower bound is
pessimistic, since not every output perturbation $\alpha$ is attainable from an input in $\mathcal{B}_p^\epsilon(X)$. Therefore, the bounds we propose in this paper do not account for local feasibility constraints in the neighborhood of $X$ (as previously discussed in Section 3.2,  Eq. 1 of \citet{schuchardtlocalized}). 
However, using this assumption allows our method to operate with negligible computational overhead, as opposed to concurrent computationally intensive approaches.

\subsection{Application to the Pixel Accuracy Metric}
\label{sec:example_pacc}

For a subset $S\subseteq \Omega$, we define the pixel accuracy on $S$ by $h: \mathcal{Y} \times \mathcal{K}^{|S|} \rightarrow \reals$ as:

\begin{equation}
    h(f(X), Y) = \frac{1}{|S|}\sum_{\omega\in S}\mathbbm{1}_{\hat{Y}_\omega=Y_\omega}
    \label{eq:pa}
\end{equation}

\noindent with $\hat{Y}_\omega$ the decision at the pixel $\omega$. Note that $S=\Omega$ is the classical pixel accuracy measure on the full image.
We define the Robust Pixel Accuracy (RPA) $h_\epsilon (X) = \min_{\tilde{X} \in \mathcal{B}_\epsilon(X)} h(f(\tilde{X}),Y)$ as the exact worst-case pixel accuracy under attack. In practice, following Eq.~\eqref{eq:min_perf}, we consider the Certifiably Robust Pixel Accuracy $\mathrm{CRPA}_\epsilon(X)$  which is a lower bound to RPA. 

\paragraph{Deriving an answer to \textbf{Q1}}
Here, we aim to answer the following question:
\textit{What is the maximal degradation of pixel accuracy that can be achieved given an adversarial budget $\epsilon$?}

\noindent We can start with the Lipschitz certificates introduced in Proposition~\ref{prop:classification-certificate}. Indeed, we recover a certificate for any particular pixel of coordinate $\omega \in \Omega$:

\begin{equation}
    R^\omega(X, Y) := \mathbbm{1}_{\hat{Y}_\omega=Y_\omega} . 2^{\frac{1-p}{p}}.\mathcal{M}_X^\omega(f) / L,
\label{eq:pixel_robustness}
\end{equation}

\noindent with $\mathcal{M}_X^\omega(f) = f^\mathrm{top1}(X)_\omega - f^\mathrm{top2}(X)_\omega$. Thus, $L.R^\omega(X, Y)$ represents the minimal norm of the perturbation $\alpha_\omega$ on the logits of pixel $\omega$ such that $f(X)_\omega+\alpha_\omega$ is misclassified. Note that $R^\omega(X, Y)$ is equal to zero for pixels already misclassified by $f(X)$. 

We propose to reformulate the $\mathrm{CRPA}_\epsilon(X)$ computation as a Knapsack Problem~\cite{salkin1975knapsack}. Let $p_\omega$ be a binary variable that equals $1$ when pixel $\omega$ is in the set of misclassified pixels under attack. We consider the evaluation of the maximum number of perturbed pixel under constraint: 
\begin{equation}
\begin{aligned}
    N_\mathrm{PA}(X,\epsilon)  = &\max  \sum_{\omega\in S} p_\omega \\
    & \;\text{s.t.}\; \sum_{\omega\in S}  L^p c_\omega p_\omega \leq (L\epsilon)^p
\end{aligned}
\label{eq:pa_knapsack}
\end{equation}

\noindent with $c_\omega = R^\omega(X, Y)^p$ the cost for a given pixel $\omega$. Here, the linear constraint $\sum_{\omega\in \Omega}  L^p c_\omega p_\omega \leq (L\epsilon)^p$ corresponds to the condition $\alpha \in \mathcal{B}_p^{L\epsilon}(0)$ in \cref{eq:min_perf} coupled with the disjoint support of output perturbations that induce pixel misclassifications. 
We can thus derive the following lower bound for the certification of \textbf{Q1}: 

\begin{equation}
    \mathrm{CRPA}_\epsilon(X) = 1 -  \frac{N_\mathrm{PA}(X, \epsilon)}{| S |}.
    \label{eq:crpa}
\end{equation}

The unidimensional Knapsack Problem described in \cref{eq:pa_knapsack} can be solved optimally in $\mathcal{O}(|S|)$ time by sorting pixels by ascending contribution $c_\omega$ and greedily adding them until the capacity is exhausted~\cite{salkin1975knapsack}.
This corresponds to the selection of less robust pixels first (already misclassified pixel having a null robustness).

By denoting $\pi_X : |S| \rightarrow S$ as a map that sorts pixel coordinates of $X$ in ascending order of robustness according to the value of $R^\omega (X,Y)$, we define:
\begin{equation}
\begin{aligned}
& N_\mathrm{SUP}(X,\epsilon, S, R^\omega) = \\
& \sup \left\{
n \in \mathbb{N}
\ \middle|\ \sum_{k=1}^{n} R^{\pi_X(k)}(X,Y)^{p} \leq \epsilon^{p}\right\}.
\label{eq:max_number}
\end{aligned}
\end{equation}

The solution of the Knapsack problem in \cref{eq:pa_knapsack}, is given by $N_\mathrm{PA}(X,\epsilon) = N_\mathrm{SUP}(X,\epsilon, S, R^\omega)$
This estimation process can be efficiently parallelized and run on GPU architectures which results in negligible overhead for the certification process compared to the forward pass of a DNN. 

\paragraph{Deriving an answer to \textbf{Q2}}
We now seek to answer the question: 
\textit{What is the maximum attack level $\epsilon$ under which the pixel accuracy is guaranteed to remain above or equal to $\gamma$?}

First, we can translate the question into a degradation criterion as defined in our framework using the previously defined pixel accuracy measure $h$:

\begin{equation}
\begin{aligned}
   \kappa [ h(f(\tilde{X}),Y) ] = 
    \mathbbm{1}_{  h(f(\tilde{X}),Y) \leq \gamma }.
\end{aligned}
\end{equation}

This corresponds to assessing the minimum norm perturbation under the constraint that at least $ \lceil \gamma . | S | \rceil $ pixels are misclassified. We define $n_\gamma:=\lceil \gamma . | S | \rceil$. Once more, this can be formulated as a Knapsack Problem\footnote{the reformulation as a maximization problem is classical by setting $q_\omega = 1 - p_\omega$.}:

\begin{equation}
\begin{aligned}
    R_\kappa(X,Y)^p \geq & \min  \sum_{\omega\in S} c_\omega p_\omega \\
    &\;\text{s.t.}\; \sum_{\omega\in S}  p_\omega \geq  n_\gamma.
\end{aligned}
\label{eq:gen_rob_ap}
\end{equation}

\noindent As for \textbf{Q1}, the solution of \cref{eq:gen_rob_ap} can be solved using the sorted pixels by:

\begin{equation}
    \underline{R}_\kappa(X,Y,S,R^\omega,n_\gamma)^p =  \sum_{k=1}^{n_\gamma} R^{\pi_X(k)} (X,Y)^p.
\label{eq:max-degradation}
\end{equation}

$\underline{R}_\kappa$ defines the  Generalized Robustness lower-bound radius for the $\gamma$ pixel accuracy over a subset 
$S\subseteq \Omega$
of the pixels of image $X$.
Note that if the clean accuracy is inferior to the threshold $\gamma$, the solution $\underline{R}_\kappa=0$, since $c_\omega = 0$ for all misclassified pixels. 

\begin{figure}
    \centering
    \includegraphics[width=0.85\linewidth]{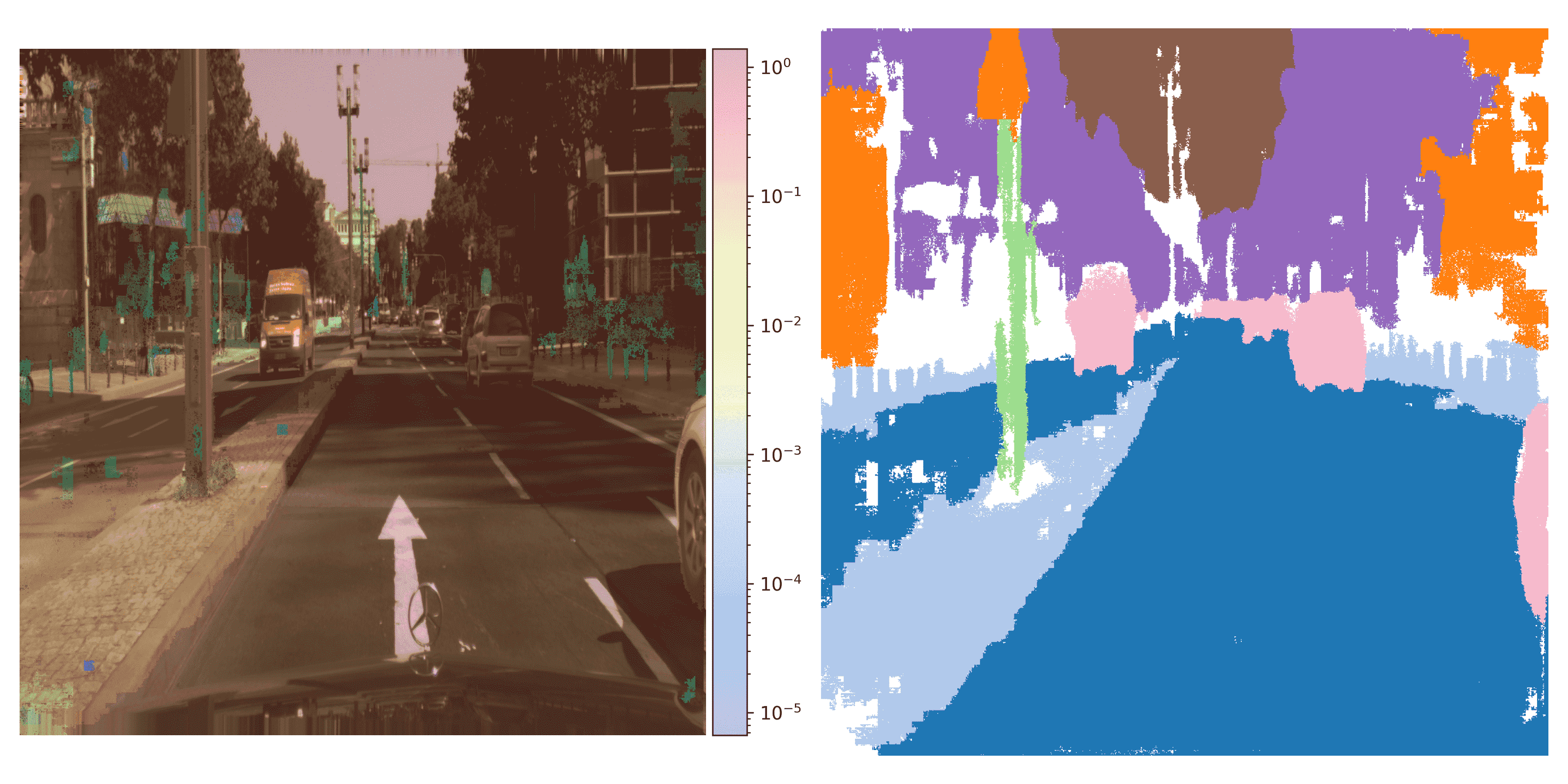}
    \caption{(Left) The $\epsilon$ budget required to attack dense segmentations to make all but $N_\mathrm{min}$ pixels change. (Right) We display only the groups of predictions where $\epsilon \geq 0.1$, non-robust pixel groups are in white.}
    \label{fig:viz}
\vspace{-10pt}
\end{figure}

\subsection{Application to FNR and Stability Metrics}
\label{sec:other_metrics}

In this section, we consider the certification of other prediction-based performance robustness measures.

\paragraph{FNR} To compute the robustness certificates for the FNR measure on a binary semantic segmentation task, we consider the subset 
$S_1  = \{\omega\in\Omega, Y_\omega = 1\}$, 
and the performance measure

\begin{equation}
\begin{aligned}
     h_\mathrm{FNR} &= \frac{1}{|S_1|}\sum_{\omega\in S_1}\mathbbm{1}_{\hat{Y}_\omega=Y_\omega} = 1-\mathrm{FNR} (\hat{Y}, Y).
\label{eq:h-fnr}
\end{aligned}
\end{equation}

We can derive an answer to \textbf{Q1} for FNR: \textit{``Which maximum level of FNR can we attain under bounded noise level $\epsilon$?''}. Similarly to PA, we can compute the maximum number of misclassified pixels in $S_1$ under an $\epsilon$ budget by 

\begin{equation} \label{eq:N(x,e)_FNR}
 N_\mathrm{FNR}(X,\epsilon)  = N_\mathrm{SUP}(X,\epsilon,S_1,R^\omega) 
\end{equation}

This provides a lower bound: 
\[
\mathrm{FNR} (\hat{Y}, Y) \leq N_\mathrm{FNR}(X,\epsilon)/|S_1|
\]

Considering the question \textbf{Q2}, \textit{``What minimum perturbation level is required to achieve a FNR above $\gamma$?''}, we consider the degradation criterion 

\begin{equation}
\begin{aligned}
   \kappa_\mathrm{FNR} [ h(f(\tilde{X}),Y)] = 
    \mathbbm{1}_{\{  h(f(\tilde{X}),Y) \leq 1-\gamma \}}.
\end{aligned}
\label{eq:kappa-fnr}
\end{equation}

We can derive a threshold index $n_\gamma=\gamma . |S_1|$ over the less robust pixels in $S_1$. 
We can use \cref{eq:max-degradation}, and obtain: 

\begin{equation}
    R_\kappa(X,Y) \geq \underline{R}_\kappa(X,Y,S_1,R^\omega,n_\gamma)
\end{equation}
which is also an easily computable lower bound for the $R_\kappa$ answering \textbf{Q2} with a degradation criterion $\kappa$ that aims to drive the FNR above a certain threshold $\gamma$.

\paragraph{Stability} Importantly, both PA and FNR require the ground-truth $Y$ values. At inference, we propose the stability measure $h_\mathrm{stab}$ on subsets of pixels $S \subseteq \Omega$ to ensure their robustness independently from the ground truth.

\begin{equation}
\begin{aligned}
     h_\mathrm{stab}(f(X),\hat{Y}^*) &= \frac{1}{|S|}\sum_{\omega\in S}\mathbbm{1}_{\hat{Y}_\omega=\hat{Y}^*_\omega}.
\label{eq:h-stab}
\end{aligned}
\end{equation}
The only difference with the pixel accuracy defined in \cref{eq:pa}, is that the reference is the clean decision $\hat{Y}^*$ when output perturbation $\alpha = 0$ (in \cref{{eq:min_perf}}).
Similarly to \cref{eq:pixel_robustness}, we denote the stability certificate for any particular pixel of coordinate $\omega \in \Omega$ (the factor $\mathbbm{1}_{\hat{Y}_\omega=\hat Y^*_\omega}$ is omitted since it is ensured by $top_1=\hat Y^*$).

\begin{equation}
    R_\mathrm{stab}^\omega(X, Y) := 2^{\frac{1-p}{p}}.\mathcal{M}_X^\omega(f) / L.
\label{eq:stability_robustness}
\end{equation}
We can now provide answers to \textbf{Q1} and \textbf{Q2} regarding the stability performance measure.
The Certified Robustness Stability is defined as
 $\mathrm{CRS}_{\epsilon}(X) = 1-N_{stab}(X,\epsilon)/|S|$
where 
\begin{equation} \label{eq:N(x,e)_stab}
\begin{aligned}
&  N_\mathrm{stab}(X,\epsilon) = N_\mathrm{SUP}(X,\epsilon,S,R_{stab}^\omega).
\end{aligned}
\end{equation}
We can also address \textbf{Q2}, which concerns the generalized stability radius for a given threshold 
$\gamma$. Setting $n_\gamma =  \lceil \gamma . | S | \rceil $, we obtain the following lower bound:
\begin{equation}
    R_\kappa(X,Y) \geq \underline{R}_\kappa(X,Y^*,S,R^\omega_{stab},n_\gamma).
\end{equation}

As an illustration, we provide a visualization of robust segmentations on the Cityscapes dataset in \cref{fig:viz} where we ensure the stability of dense subsets of predictions up to a radius $\epsilon$ such that all but a fixed number of pixels (set by the user) of said objects could change class under attack. 
Additionally, we also provide robustness certificates for more complex performance measures such as the class IoU. We detail our computations in Appendix B.

    \section{Experimental Validation}

In this section, we provide more context on Lipschitz network training and apply our methods to certify various objectives. We compare our methods certificates against the \textsc{SegCertify} method, and evaluate the tightness of Lipschitz bounds by using adversarial attacks. Finally, we demonstrate the flexibility of Lipschitz-based certification in a safety-critical scenario.

\subsection{Lipschitz by design networks}

The exact computation of a network’s Lipschitz constant (see Def.~\ref{def:lipschitz}) is known to be NP-hard~\cite{virmaux2018lipschitz}. As a practical alternative, Anil et al.~\cite{anil2019sorting} introduced architectures in which every linear layer is constrained such that $\| \nabla_x f(x) \|_2 = 1$ almost everywhere. As a result, the multiplicative bound $L_f = \prod_{i=1}^L L_{f_i}$ of layer wise Lipschitz constants $L_{f_i}$, becomes a meaningful certificate on the network’s global Lipschitz constant.

Importantly, most of these orthogonality constraints are ensured by differentiable re-parametrization on the neural networks weights, which can be performed efficiently, resulting in networks with limited training overhead w.r.t their standard unconstrained counterparts (see Appendix E).
Moreover, constraining the Lipschitz constant of a neural network exhibits several other benefits, such as explicit control of the network's position on the accuracy-robustness tradeoff \cite{bethune2022pay}, or even ensuring training stability~\cite{bansal2018gain,newhouse2025training} for state-of-the-art deep learning tasks. 

We use DeepLabV3 \cite{chen2017deeplab} as our baseline architecture for semantic segmentation, since, to the best of our knowledge, it achieves state-of-the-art performance among convolutional neural network (CNN)–based methods. Additionally, this architecture is perfectly compatible with real time inference.
Attention-based architectures were excluded 
since standard attention layers are not Lipschitz-continuous, and only very recent works have started to address this limitation~\cite{kim2021lipschitz}. 
DeepLabV3 employs dilated convolutions and pooling operations to capture multi-scale contextual information. To design a Lipschitz-constrained version of DeepLabV3, we leverage the \textsc{orthogonium} and \textsc{deel-lip} libraries introduced in \citet{boissin2025adaptive} and \citet{serrurier2021achieving} respectively. 
More details on this network architecture are provided in Appendix D.

\begin{figure}
    \centering
    \includegraphics[width=0.90\linewidth]{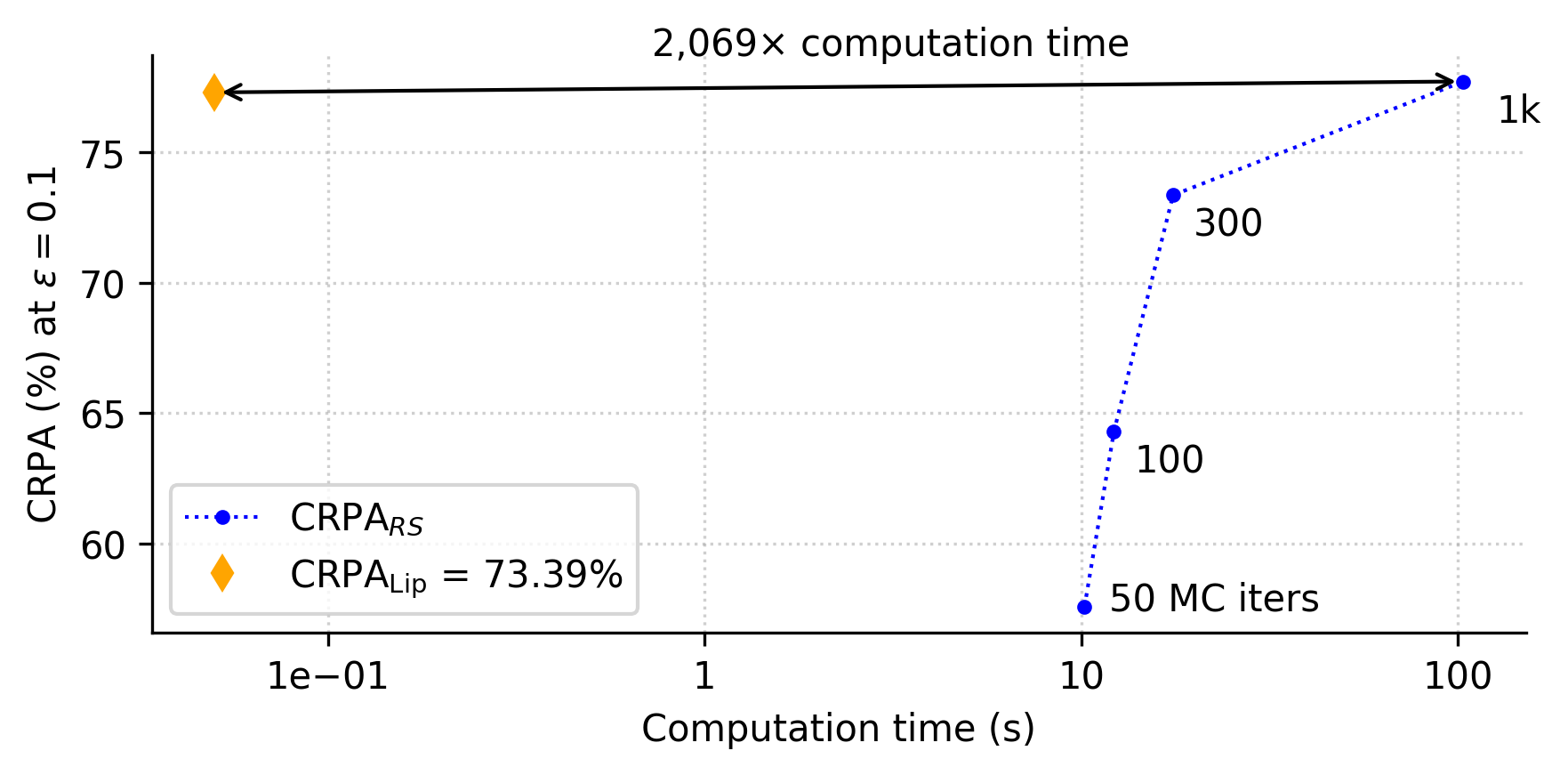}
    \caption{We evaluate the certified pixel accuracy of \textsc{SegCertify}
    on the same $L$-Lipschitz neural network that performs Semantic Segmentation on the Oxford-IIIT Pet dataset~\cite{parkhi2012cats}. For \textsc{SegCertify} we carefully select $\sigma$ from $\{0.035, 0.05, 0.08, 0.1, 0.15, 0.2, 0.3 \}$ for each number of MC samples. Here, we certify the pixel accuracy against $\epsilon=0.1$ in $\ell_2$ norm. We pick $\alpha=0.01$ as the failure probability for $\mathrm{CRPA}_\mathrm{RS}$. }
    \label{fig:lipschitz_mc}
\vspace{-10pt}
\end{figure}

\begin{table*}[t]
\centering
\renewcommand{\arraystretch}{1.2}
\setlength{\tabcolsep}{10pt}
\begin{tabular}{c l c c c}
\toprule
$\epsilon$ & \textbf{Method} & \textbf{CRPA} & \textbf{Time (total / nb samples)} & \textbf{$\#$ forward passes / sample} \\
\midrule
    0.1 & Lipschitz bound (ours) & $81.80\%$ & $\approx 0.1$ s & 1 \\
    0.1 & \textsc{SegCertify} ($\sigma=0.3$)   & $53.48 \pm 0.59 \% $ & 59.8 s \color{red}{\small{$\times$594}} & 60 \\
    0.1 & \textsc{SegCertify} ($\sigma=0.2$)  & $83.13 \pm 0.33 \% $ & 62.1 s \color{red}{\small{$\times$624}} & 80 \\
    \midrule
    0.17 & Lipschitz bound (ours) & $77.34\% $  & $\approx 0.1$ s & 1 \\
    0.17 & \textsc{SegCertify} ($\sigma=0.4$)   & $38.91 \pm 0.53 \%$  & 60.3 s \color{red}{\small{$\times$594}} & 60 \\
    0.17 & \textsc{SegCertify} ($\sigma=0.2$)   & $84.84 \pm 0.73 \%$  & 63.3 s \color{red}{\small{$\times$683}} & 120 \\
\bottomrule
\end{tabular}
\caption{CRPA values across methods on the Cityscapes dataset \cite{cordts2016cityscapes} using $1024 \times 1024$ images. We choose $\alpha = 0.001$ as the failure probability of \textsc{SegCertify} and tune $\sigma \in \{0.15, 0.2, 0.25, 0.3, 0.4, 0.5\}$ for each run. Finally, given the very long computation time of smoothing based methods, evaluations are run on only $100$ images of the dataset, as done in \cite{fischer2021scalable}. We report the mean and standard deviation of results across 5 runs that use the best performing $\sigma$ value. We also report the mean runtime for each evaluation divided by the number of samples. The results using Lipschitz bounds are obtained on the whole test set.}
\label{tab:cert_results}
\vspace{-12pt}
\end{table*}

\subsection{Comparing LipNet certificates to SegCertify}
\label{sec:comparing_methods}

In order to compare Randomized Smoothing based approaches with our proposed Lipschitz approach, we must define a shared performance metric. 
In the setting of Lipschitz constrained networks, we are able to directly compute:

\begin{equation}
    \mathrm{CRPA}_\mathrm{Lip} = 1 - \frac{N(X, \epsilon)}{| \Omega |},
\end{equation}
In order to enable fair comparison between methods, we should penalize abstentions as they still impede the decision making process of downstream systems that might depend on the segmentation network. We adopt the following metric to compare Lipschitz and Randomized Smoothing based certificates:
 
\begin{equation}
    \mathrm{CRPA}_\mathrm{RS} := \frac{1}{| \Omega |} \times \sum_{\omega \in \Omega} \mathbbm{1}_{\hat Y_\omega \neq \emptyset \wedge \hat Y_\omega = Y_\omega}
\end{equation}
which recovers the standard pixel accuracy formulation with abstentions considered as misclassifications.

\subsubsection{Using Different Networks}

Using the common comparison metric we defined in Section~\ref{sec:comparing_methods}, we compare a Lipschitz constrained DeepLabV3 architecture to its unconstrained counterpart on which we run the randomized smoothing method of~\citet{fischer2021scalable}. We train our unconstrained model with a noise level that is similar to the $\sigma$ value we will choose for the smoothing process according to~\citet{salman2019provably}. The CRPA certificates given by both Lipschitz constraints and the \textsc{SegCertify} method are presented in Table~\ref{tab:cert_results}. We also report the time each evaluation took divided by the number of evaluated samples. It does not scale linearly in MC iterations since the computation time of \textsc{SegCertify} depends on both the MC estimation runtime (which can be partially batched, but not fully due to memory limits) and the $p$-value computation times which are dependent of the image size. More experimental details are given in Appendix F.

We observe that Lipschitz-constrained networks allow for blazingly fast inference and certification with performance that remains competitive with that of randomized smoothing methods. Crucially, the performance obtained by using a LipNet is only matched by randomized smoothing methods when they require approximately 600 times more time per inference.

\subsubsection{Comparing CRPAs on the Same LipNet}

We might wonder how tight our Lipschitz bounds are, and whether the better performance of Randomized Smoothing methods for higher computational budgets at inference (e.g. for 120 forward passes at $\epsilon=0.17$) stems from tighter approximations of the local Lipschitz constant or just better base model accuracy of the unconstrained model.

In order to answer this question, we devise the following experiment: on the same $L$-Lipschitz DeepLabV3 network, we compare the CRPA certificate of the \textsc{SegCertify} method with a varying number of MC samples with the CRPA certificate provided by our Lipschitz bound. Indeed, since increasing the number of MC samples improves the performance of randomized smoothing methods this will allow us to determine the minimum number of MC samples that is necessary for probabilistic methods to match our deterministic bound on the same base model. Our results are reported in Fig.~\ref{fig:lipschitz_mc}. These results show that randomized smoothing methods need more that $1000$ MC samples to become beneficial over using a Lipschitz certificate. On the IIIT Pets dataset with $128 \times 128$ images,  equates to an $\approx 2000$ times slower runtime per image. Thus the $\mathrm{CRPA}_\mathrm{Lip}$ method evaluated on the same neural network is more than 2000 times faster at equal robustness levels.

\subsection{About the Empirical Robustness of LipNets}

\begin{figure}
    \centering
    \includegraphics[width=0.9\linewidth]{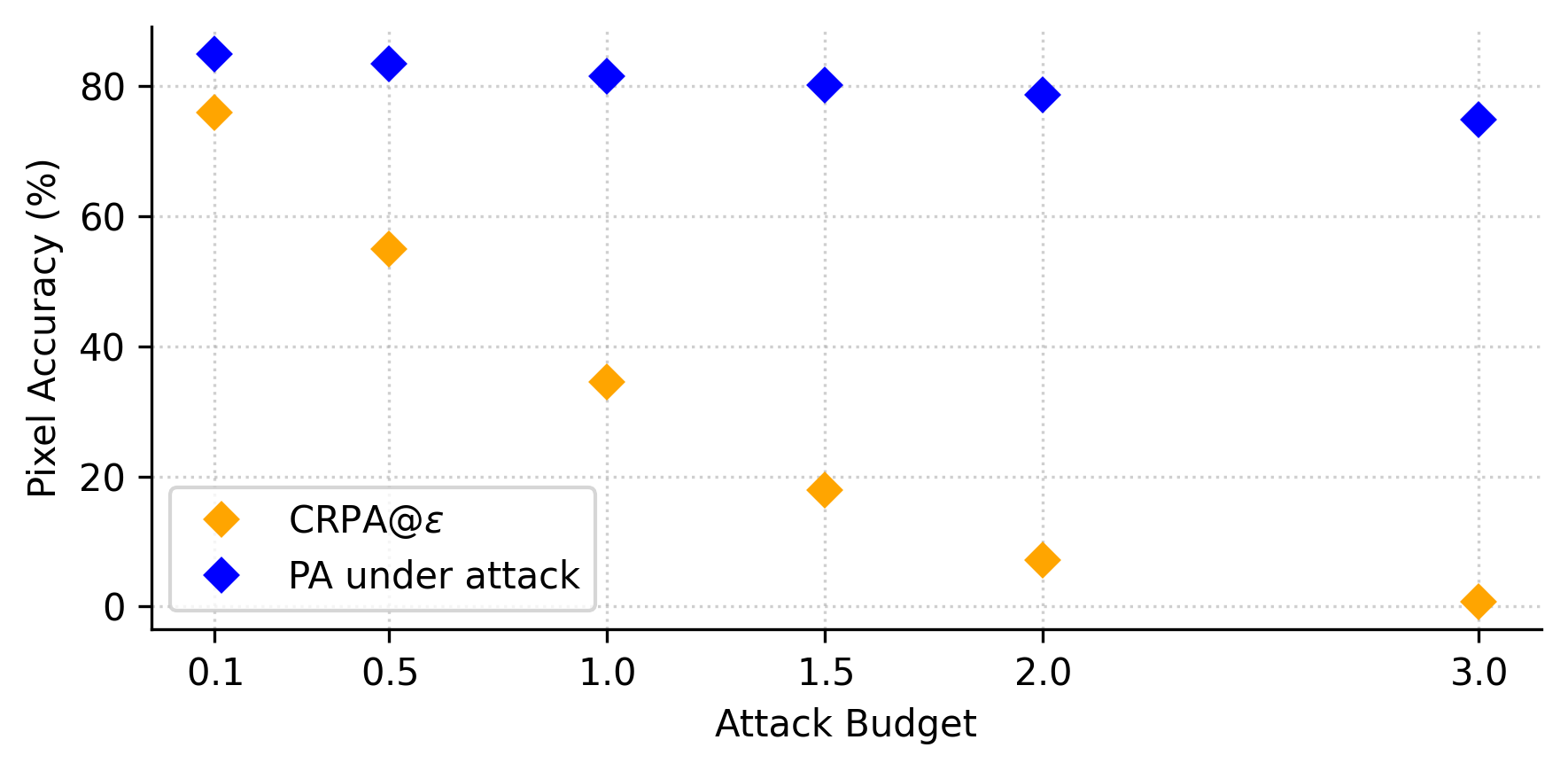}
    \caption{Here, we attack 500 images from the Oxford-IIIT Pet dataset, for each different adversarial budget, we plot the CRPA and the actual empirical pixel accuracy under attack.}
    \label{fig:attacks}
    \vspace{-15pt}
\end{figure}

In order to obtain an upper-bound for the worst-case performance of Semantic Segmentation networks under $\epsilon$ bounded adversarial attacks, we apply segmentation specific adversarial attacks from various different works on the Oxford-IIIT Pet dataset. 
Here, we use the ALMA, ASMA and PDPGD attacks from \cite{rony2021augmented,xie2017adversarial,matyasko2021pdpgd} respectively, as implemented in~\cite{Rony_Adversarial_Library}.
For every image, we choose the best-performing attack from the previously mentioned subset of methods and compile the overall worse pixel accuracy we could get from attacking the test images. As shown in~\cref{fig:attacks}, our networks are very robust to adversarial perturbations. Moreover, the gap between certificates and empirical instantiations seems to grow larger when $\epsilon$ gets bigger. The reason behind this is threefold: Firstly, our Lipschitz bounds are tight in low $\epsilon$ settings, and ensuring robustness in higher $\epsilon$ regimes requires more robust but less accurate networks. Secondly, as $\epsilon$ grows, output perturbations described in Eq.~\ref{eq:min_perf} become less and less feasible. Finally, empirical adversarial attacks only provide an upper-bound for the worst-case performance under attack which is be overly optimistic regarding the worst-case adversarial robustness.

\subsection{A Safety Critical Use-Case}
\label{sec:q2}

In this section, we will further demonstrate how Lipschitz-constrained networks allow for fast and practical worst-case bound in Safety-Critical scenarios. In this setting, we will be interested in a binary segmentation task that handles the detection of polyps from endoscopically capture images. To this end, we use the Kvasir-SEG dataset introduced in~\citet{jha2020kvasir}. Typically, this is a safety-critical scenario as misdetections might lead to harmful consequences on the patient. Importantly, the deployment of such a detection model must satisfy some real-time constraints given the time-sensitive nature of the assisted surgeon's task. 

\begin{table}[t]
\centering
\label{tab:q1q2}
\begin{tabular}
{@{}S[table-format=1.3] S[table-format=1.3] S[table-format=1.3] S[table-format=1.3]@{}}
\toprule
\multicolumn{2}{c}{\textbf{\textbf{Q1}}} & \multicolumn{2}{c}{\textbf{\textbf{Q2}}} \\
\cmidrule(lr){1-2}\cmidrule(lr){3-4}
\multicolumn{1}{c}{$\epsilon$} & \multicolumn{1}{c}{Certified FNR} &
\multicolumn{1}{c}{$\gamma$} & \multicolumn{1}{c}{Required $\epsilon$} \\
\midrule
0.1  & 0.612 & 0.7  & 0.423 \\
0.2  & 0.768 & 0.85 & 0.563 \\
0.3  & 0.871 & 0.95 & 0.675 \\
\bottomrule
\end{tabular}
\caption{Here, we present the lower-bounds obtained as answers to \textbf{Q1} and \textbf{Q2} type certification processes. 
}
\label{tab:fnr}
\vspace{-10pt}
\end{table}

\paragraph{Defining meaningful robustness certificates} Here, we want to certify the worst-case False Negative Rate of our segmentation network under adversarial noise bounded by $\epsilon$. In the following, we use $h$ and $\kappa$ defined as in Equation~\ref{eq:h-fnr}~and~\ref{eq:kappa-fnr} respectively. We expose our robustness certificates in the setting of both \textbf{Q1} and \textbf{Q2} in Table~\ref{tab:fnr}. 
Our \textbf{Q1} certificates can be read as ``given budget $\epsilon$, I cannot attack images so that $\mathrm{FNR} \geq \gamma$''. For \textbf{Q2}, we can understand ``To ensure $\mathrm{FNR} \geq \gamma$ on the test set, I need a mean $\epsilon$ budget across test images''.
Most importantly, the forward pass of our Lipschitz segmentation network only requires $\approx 0.05$ s which ensures compatibility with real-time applications.

    \section{Perspectives and Conclusion}

In this paper, we propose Lipschitz neural networks for semantic segmentation tasks, which enables real-time compatible certifiably robust segmentation for the first time. Also, we show how to train and use Lipschitz neural networks to obtain general robustness certificates efficiently in a variety of different scenarios.

\paragraph{Limitations and Future Research Directions} While Lipschitz networks allow for an efficient certification process for robustness certificates on semantic segmentation predictions. They struggle to match the performance of smoothed state-of-the-art semantic segmentation networks that leverage a high number of MC iterations or even diffusion models~\cite{laousy2023certification}.
Indeed, if we consider the performance to inference time compute Pareto frontier, we have that Lipschitz constrained neural networks are at one end of the spectrum, whereas randomized smoothing based methods are on the other end, offering impressive performance in high compute scenarios. Future work could focus on developing hybrid alternatives that leverage model architecture and input dependencies (such as receptive fields\cite{luo2016understanding, schuchardtlocalized}), in order to make the bound of Equation~\ref{eq:min_perf} tighter by incorporating feasibility constraints into \cref{eq:min_perf}, while keeping real-time compatibility. 
Also, adding considerations on the distributions of inputs could be useful to allow fast and meaningful certificates~\cite{zargarbashi2025one,hashemi2025probabilistic}.

\section{Acknowledgements}

This work was carried out within the DEEL project,\footnote{\url{https://www.deel.ai/}} which is part of IRT Saint Exupéry and the ANITI AI cluster. The authors acknowledge the financial support from DEEL's Industrial and Academic Members and the France 2030 program – Grant agreements n°ANR-10-AIRT-01 and n°ANR-23-IACL-0002. 
This work was granted access to the HPC resources of IDRIS under the allocation 2025-AD011016850 made by GENCI. 

\hfill

Additionally, the authors thank Thibaut Boissin for his valuable input regarding figures and phrasing, as well as Tom Rousseau and Laurent Gard\`es for their guidance and continued support of the project.
\else

\fi 

{
    \small
    \bibliographystyle{ieeenat_fullname}
    \bibliography{main}
}

\appendix

\onecolumn

\section{Proof}
\label{app:proof}

In this section, we provide some theoretical context on the certification processes described in Section~\ref{sec:other_metrics} of the paper.

\subsection{Proof of Equation~\ref{eq:min_perf}}
\label{sec:proof_min_perf}

For sake of readibility, we copy here the equation, given $f$ an $L$-Lipschitz function with respect to the $\ell_p$ norm, we have:

\begin{equation}
\begin{aligned}
     h_\epsilon (X) &= \min_{\tilde{X} \in \mathcal{B}_p^{\epsilon}(X)} h(f(\tilde{X}), Y) \\
     &\geq \min_{\alpha \in \mathcal{B}_p^{L\epsilon}(0)} h (f(X) + \alpha,Y).
\end{aligned}
\end{equation}

\begin{proof}
By definition,
\[
    h_\epsilon(X) = \min_{\tilde{X} \in \mathcal{B}_p^{\epsilon}(X)} h(f(\tilde{X}), Y),
\]
where $\mathcal{B}_p^{\epsilon}(X) = \{ \tilde{X} \in \mathcal{X} : \| \tilde{X} - X \|_p \leq \epsilon \}$.

For any $\tilde{X} \in \mathcal{B}_p^{\epsilon}(X)$, the Lipschitz condition on $f$ implies
\[
    \| f(\tilde{X}) - f(X) \|_p \leq L \| \tilde{X} - X \|_p \leq L \epsilon.
\]
Thus, for each such $\tilde{X}$, $\alpha = f(\tilde{X}) - f(X)$ satisfies $\| \alpha \|_p \leq L \epsilon$.
In other words, $f(\tilde{X}) \in f(X) + \mathcal{B}_p^{L\epsilon}(0)$.

Therefore, we can write
\[
    \{ f(\tilde{X}) : \tilde{X} \in \mathcal{B}_p^{\epsilon}(X) \} \subseteq f(X) + \mathcal{B}_p^{L\epsilon}(0).
\]
Applying the monotonicity of the minimum over subsets, we get
\[
    \min_{\tilde{X} \in \mathcal{B}_p^{\epsilon}(X)} h(f(\tilde{X}), Y)
    \geq \min_{\alpha \in \mathcal{B}_p^{L\epsilon}(0)} h(f(X) + \alpha, Y).
\]
Hence,
\[
    h_\epsilon (X)
    \geq \min_{\alpha \in \mathcal{B}_p^{L\epsilon}(0)} h(f(X) + \alpha, Y),
\]
which completes the proof.
\end{proof}

\subsection{About Equation~\ref{eq:max_number}}

Formally, the knapsack problem can be defined as follows: We are given an instance of the knapsack problem with item set $\Omega$, consisting of \textit{$n$ items $\omega$} with profit $p_\omega$ and weight $c_\omega$, and the capacity value $C$. Then, the objective is to select a subset of $\Omega$ such that the total profit of the selected items is maximized and the total weight does not exceed $C$, as per~\citet{kellerer2004multidimensional}.

\hfill

In our case, \cref{eq:max_number} is obtained via the simplest case of knapsack instances, where for each pixel coordinate $\omega \in \Omega$, we have $p_\omega = 1$ if we attack the correctly classified pixel $\omega$ by using budget $c_\omega = R^\omega(X,Y)^p$, we aim to keep the total minimum required adversarial budget under $C = \epsilon^p$. We denote $x_\omega$ the binary variable that represents the selection or not of the object in the knapscak (i.e. if the pixel $\omega$ is attacked or not). Our knapsack formulation is to find the optimal $\{x_\omega\}_{\omega \in \Omega}$ as:

\begin{equation}
\begin{aligned}
    & \text{maximize} \ \sum_{\omega \in \Omega} p_\omega.x_\omega, \\
    & \text{subject to} \ \sum_{\omega \in \Omega} c_\omega.x_\omega \leq C, \\
    & x_\omega \in \{0, 1\}, \ \omega \in \llbracket 1, | \Omega | \rrbracket.
\end{aligned}
\end{equation}

Consequently, we have that our formulation recovers that of a classical knapsack problem. In the case of the pixel accuracy, $p_\omega = 1$ everywhere and $c_\omega = R^\omega(X,Y)^p$, which can be solved efficiently by sorting the items of the set $\Omega$ by ascending cost of $c_\omega$ and obtaining the index such that the cumulative sum of the sorted $c_{\pi(\omega)}$ is inferior to $C = \epsilon^p$. This corresponds to finding the optimal vector $x^* = (x_1^*, ..., x^*_{|\Omega|})$ where the attack that perturbs the most correctly classified pixels misclassifies all $i$ such that $x_i = 1$.

\section{Worst-case Class IoU Computation} 
\label{app:wc_iou}

In this section, we provide an explicit description of the more complex computation of robustness certificates for the class IoU computation. Indeed, we have that computing the worst-case class IoU value under $\epsilon$-bounded adversarial attacks is equivalent to a fractional knapsack problem, which requires a greedy but yet still efficient certification process. We recall the class IoU performance measure as:

\begin{equation} 
    h_{\mathrm{IoU}_k}(X,Y) := \frac{\sum_{\omega \in \Omega} \mathbbm{1}_{(\hat Y_\omega = Y_\omega) \wedge (Y_\omega = k)}}{\sum_{\omega \in \Omega} \mathbbm{1}_{(\hat Y_\omega = k) \vee (Y_\omega = k)}}.
\label{eq:class_iou}
\end{equation}

\hfill

\noindent Importantly, the class IoU can be degraded in two (non-exclusive) ways: 
\begin{itemize}
    \item changing the prediction of true positive (TP) pixels to any class other than $k$, i.e. a smaller numerator in Eq.~\eqref{eq:class_iou}. The certificate to perturb a TP pixel is $R_{TP}^\omega = f_\omega^k - f_\omega^{\mathrm{top2}}$.
    \item changing the prediction of true negative (TN) pixels to class $k$, i.e. a higher denominator in Eq.~\eqref{eq:class_iou}. The certificate to perturb a TN pixel is $R_{TN}^\omega = f_\omega^{\mathrm{top1}} - f_\omega^k$.
\end{itemize}

\hfill 

We introduce $S_k= \{\omega\in\Omega : Y_\omega = k \}$, and the binary variable $p_\omega = 1$ for the set of pixels in $S_k$ misclassified under attack, i.e. $p_\omega = (\hat Y_\omega \neq c) \wedge (Y_\omega = c)$. We also denote $q_\omega$, the set of pixels in $\Omega\setminus S_k$ that are are  classified as $k$ under attack: $q_\omega = (\hat Y_\omega = k) \wedge (Y_\omega \neq k)$.
We can derive a formulation of $h_{\mathrm{IoU}_k}$ with $p_\omega$ and $q_\omega$:

\begin{equation} 
    h_{\mathrm{IoU}_k}(X,Y) := \frac{|S_k|-\sum_{\omega \in S_k} p_\omega}{|S_k| + \sum_{\omega \in \Omega\setminus S_k} q_\omega}.
\end{equation}

We denote $c_\omega=(R_{TP}^\omega)^p$ for any $\omega\in S_k$, and $c_\omega=(R_{TN}^\omega)^p$ for any $\omega\in \Omega\setminus S_k$.
We can reformulate the lower bound of $h_{\mathrm{IoU}_k}$ as a Linear Fractional Programming problem:

\begin{equation}
\begin{aligned}
    &\min  \frac{|S_k|-\sum_{\omega \in S_k} p_\omega}{|S_k| + \sum_{\omega \in \Omega\setminus S_k} q_\omega} \\
    & \;\text{s.t.}\; \sum_{\omega\in S_k}  L^p c_\omega p_\omega + \sum_{\omega\in \Omega\setminus S_k}  L^p c_\omega q_\omega \leq (L\epsilon)^p
\end{aligned}
\label{eq:iou_lfp}
\end{equation}

Crucially, we solve this problem with a two-stage greedy strategy. For each possible amount of budget allocated to increasing the numerator (by misclassifying TP pixels in order of increasing $c_\omega$), we then greedily use the remaining budget to decrease the denominator (by misclassifying TN pixels as class $k$, again in increasing $c_\omega$). Among all such budget splits, we retain the configuration that yields the smallest IoU.
The code for this process is provided in our codebase. Importantly, this method allows the certification of the worst-case IoU of the ``\textit{pedestrian}'' class on the whole test set of the Cityscapes dataset in only 38 seconds (on $1024 \times 1024$ images), i.e 8 milliseconds per image on average. Our results our presented in \cref{tab:wc_iou}.


\begin{table}[ht]
    \centering
    \begin{tabular}{@{}cc@{}}
        \toprule
        $\epsilon$ & Worst Reachable Class IoU \\
        \midrule
         0.0 & 47.37 \\
         $1.0 \times 10^{-2}$ & 31.76 \\
         $2.5 \times 10^{-2}$ & 21.15 \\
         $5.0 \times 10^{-2}$ & 11.43 \\
         $7.5 \times 10^{-2}$ & 06.49 \\
         $1.0 \times 10^{-1}$ & 03.83 \\
        \bottomrule
    \end{tabular}
    \caption{Here, we compute a certificate for the worst case class IoU of the ``\textit{person}'' class on the Cityscapes dataset under $\epsilon$ bounded attacks using our algorithm. This is particularly difficult to certify given the small proportion of pixels that this class represents.}
\label{tab:wc_iou}
\end{table}

\begin{figure}[ht!]
    \centering
    \includegraphics[width=0.89\textwidth]{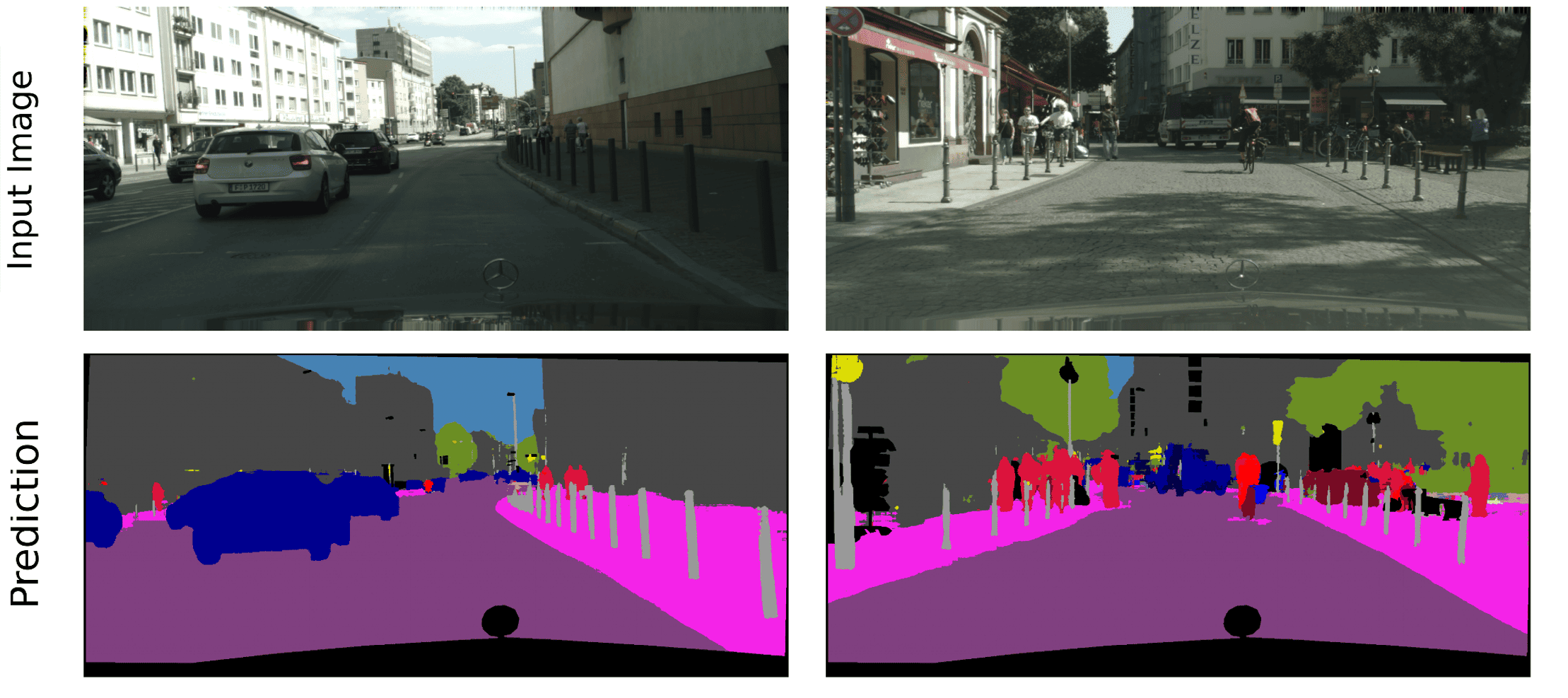}
    \caption{Visualization of test set segmentation results using our Lipschitz constrained neural networks trained using the cosine similarity.}
    \label{fig:placeholder}
\end{figure}

\section{Segmentation Specific Adaptations}

In this section, we provide more detail regarding our implementation of Lipschitz networks for segmentation tasks. 

\hfill

First, we scale the outputs of residual connections that add or concatenate the residual outputs accordingly to the Lipschitz constant of the operation being realized. We ensure the Lipschitz constant of the network remains unchanged by scaling the outputs of concatenations by $1/\sqrt{N}$ with $N$ the number of different outputs that get concatenated (and $1/N$ when the outputs are added).

\hfill

Also, to ensure that our Lipschitz network does not suffer from stability issues due to small activation values, we use the scaling method of \citet{yang2023spectral}. This scaling factor ensures that the RMS norm of feature maps stays stable and therefore allows us to replace BatchNormalization layers by BatchCentering layers which exhibit 1-Lipschitz behavior.

\hfill

To fairly characterize how expressive our DNNs architectures are we train the same DeepLabV3 architecture with both unconstrained and Lipschitz building blocks, we will refer to these models as \textit{AllNet} and \textit{LipNet} respectively. We train our Lipschitz network to maximize accuracy (i.e with a high temperature). We obtain the following results:
\begin{table}[h]
    \centering
    \begin{tabular}{@{}ccc@{}}
        \toprule
        Model & Pixel Acc. & mIoU \\
        \midrule
        LipNet & 92.07\% & 51.80 \\
        AllNet & 94.41\% & 64.55 \\
        \bottomrule
    \end{tabular}
    \caption{Comparing the expressivity of Lipschitz-constrained blocks to standard unconstrained blocks on the Cityscapes dataset.}
\end{table}

\begin{figure}[ht]
    \centering
    \includegraphics[width=0.6\linewidth]{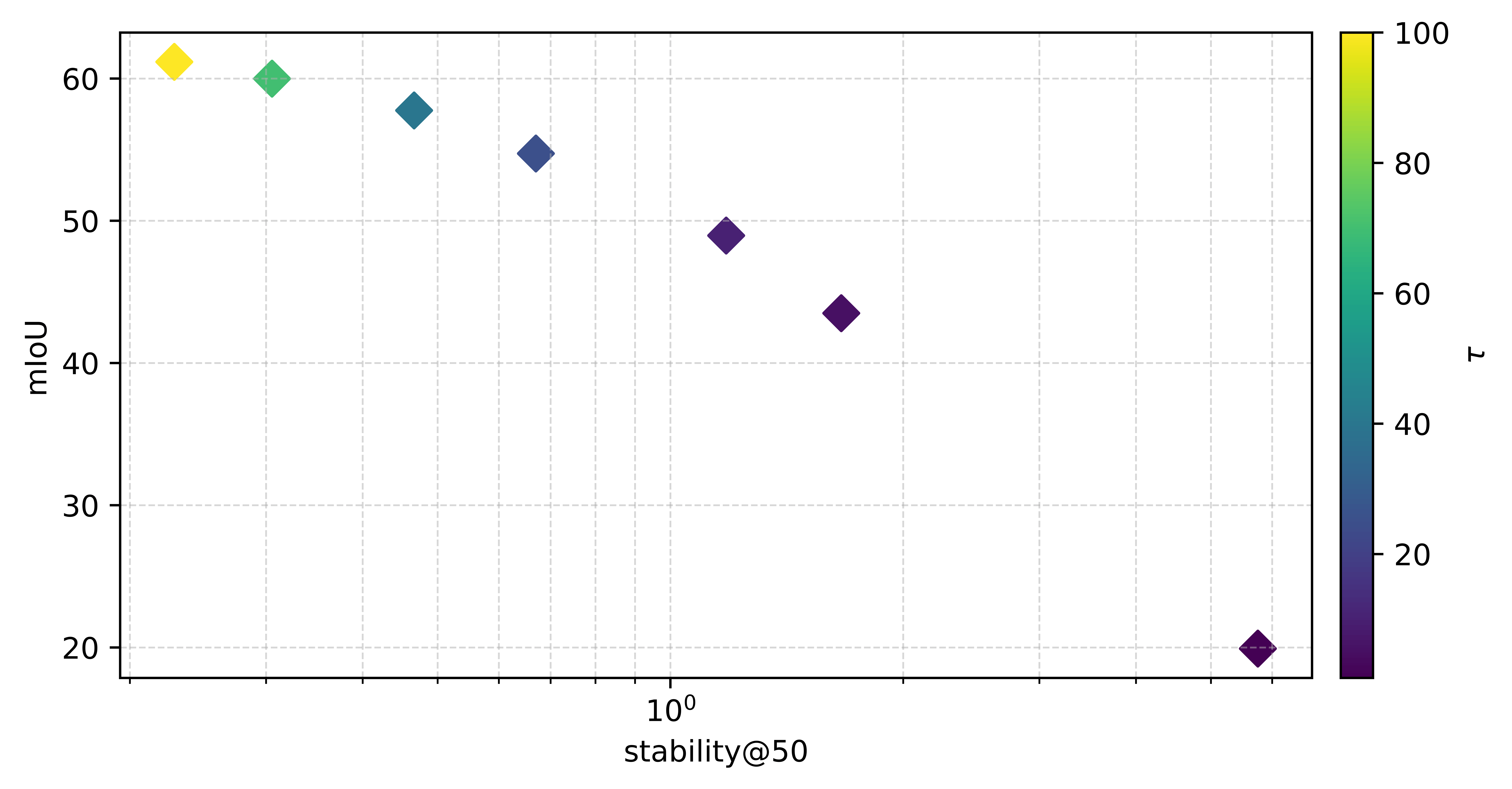}
    \caption{Lipschitz neural networks allow for explicit control of their position on the robustness-accuracy tradeoff \cite{bethune2022pay}, as seen on this small network trained on the Cats \& Dogs dataset. This property remains valid in the context of semantic segmentation.}
    \label{fig:robustnesspareto}
\end{figure}

In practice however, the decisions of this accurate LipNet model are too brittle to certify meaningful CRPA for non trivial $\epsilon$ values. In the rest of the paper, we use scaled temperature cross-entropy training to train networks that provide non vacuous robustness certificates while maintaining informative enough pixel accuracy values in a clean setting. 
As per~\citet{bethune2022pay}, we train Lipschitz neural networks with a temperature-scaled cross-entropy objective. This allows us to explicitely control the position of our trained neural networks on the accuracy-robustness tradeoff, as depicted in Fig.~\ref{fig:robustnesspareto}.

\paragraph{Characterizing a LipNet's robustness} In order to characterize the robustness of a Lipschitz neural network independently from its accuracy, we denote the ``stability'' measure as a direct measure of a neural network prediction's invariance. Its formulation is equivalent to that of Equation~\ref{eq:pixel_robustness}, except the correctness of the prediction is not accounted for.
We use this metric along with the mIoU metric to demonstrate how training Lipschitz neural networks on segmentation tasks also allows us to control the placement of our trained model on the robustness-accuracy Pareto frontier (see Fig.~\ref{fig:robustnesspareto}). Here, we denote the \texttt{stability@50} measure as a lower-bound on the minimum required adversarial budget $\epsilon$ to change at least $50\%$ of the networks pixel-wise predictions.

\section{About the DeepLabV3 architecture}
\label{app:deeplab}

\begin{figure}[h]
    \centering
    \includegraphics[width=0.95\linewidth]{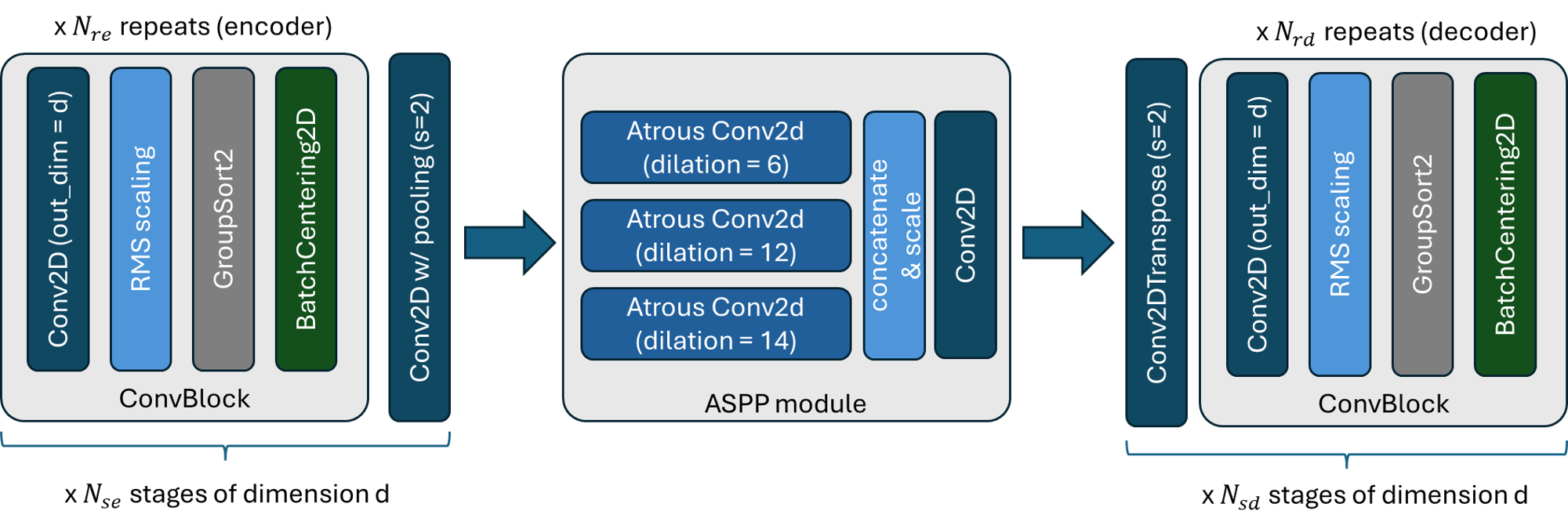}
    \caption{The DeepLabV3 architecture relies on an extraction of fine and coarser features by leveraging the atrous convolutions in the ASPP modules, which each use a different dilation to capture finer or larger features.}
    \label{fig:deeplab}
\end{figure}

In our paper, we use a DeepLabV3-like architecture across experiments. It is described in Figure~\ref{fig:deeplab}.

We provide different configurations for this architecture by changing $N_{re}$, $N_{rd}$, $d$ for each stages along with the number of stages $N_{se}$ ($N_{sd}$ is chosen to recover the original image's size). 

\paragraph{Why choose a DeepLabV3 architecture?} In order to choose a good architecture for our segmentation networks, we run an experiment on the Oxford-IIIT Pet dataset where we train a variety of different networks by varying: the parametrization (Lipschitz-constrained, orthogonally constrained or unconstrained), the loss function (Hinge Kantorovich-Rubinstein~\cite{serrurier2021achieving}, cosine similarity or scaled cross-entropy~\cite{bethune2022pay}), and finally the neural network architecture: varying between FCN~\cite{long2015fully}, UNet~\cite{ronneberger2015u} and DeepLab~\cite{chen2017deeplab}. We obtain a mean mIoU improvement of 3.3 and 1.9 points, respectively, in favor of the DeepLabV3 architecture w.r.t. the FCN and UNet architectures.

\section{Training Overhead of Lipschitz Neural Networks}
\label{app:overhead}

Training Lipschitz neural networks introduces a computational overhead that is twofold: First, the Lipschitz layers must be parametrized with an efficient differentiable parametrization method, this ensures that the Lipschitz constant of the layer is at most $1$. Secondly, the weight updates must be backpropagated through this parametrization in order to train the neural network using the chain rule. While this extra overhead might seem costly, efficient parametrizations exist in order to parametrize Lipschitz convolutive layers that allow for tight certificates~\cite{boissin2025adaptive}.

\hfill

Also, the parametrization methods Lipschitz neural networks leverage only necessitate to be recomputed after every weight update, and their cost is not input dependant. Therefore, their cost is rather limited when using larger batch sizes, making Lipschitz neural networks more efficient at scale when batch sizes are increased. To demonstrate this phenomenon, we train the same DeepLab architecture with unconstrained and Lipschitz parametrized blocks for different batch sizes on the IIIT Pets dataset. We obtain the following results:


\begin{table}[ht]
    \centering
    \begin{tabular}{cccc}
        \midrule
        Network & Batch size & Training speed (batches/s) & Runtime overhead (vs AllNet) \\
        \midrule
        AllNet & 30  & 12.58 & 1.00$\times$ \\
        LipNet & 30  & 5.33 & 2.36$\times$ \\
        \midrule
        AllNet & 100 &  3.94 & 1.00$\times$ \\
        LipNet & 100 &  3.62 & 1.09$\times$ \\
        \midrule
    \end{tabular}
    \caption{Number of gradient steps per second for different batch sizes. We use the M1 config of our DeepLabV3 architecture on the Oxford-IIIT Pet dataset with $128 \times 128$ images and the parametrization from \cite{boissin2025adaptive}. The last column reports the relative training speed of the Lipschitz-constrained LipNet compared to the unconstrained AllNet (higher is slower, AllNet $=1.00\times$).}
    \label{tab:speed}
\end{table}


Therefore, while Lipschitz neural networks introduce a computational overhead at train time. That overhead becomes rather negligible in large batch size regimes. Moreover, offloading the parametrization cost to a single GPU can be beneficial in multi-GPU settings, further reducing the overhead of Lipschitz parametrizations.

\section{Experimental Settings}
\label{app:experiments}

During our training runs, we use the following settings:

\paragraph{Input sizes} We use the following image sizes on these datasets:
\begin{itemize}
    \item Oxford-IIIT Pet: $128 \times 128$.
    \item Cityscapes: $1024 \times 1024$.
    \item Kvasir-SEG: $256 \times 256$.
\end{itemize}

\paragraph{Model configuration} In our configuration, we propose the following configuration for our DeepLabV3 configurations: 
\begin{itemize}
    \item \textbf{S}: which uses $N_{re} = 3$ with three stages of dimensions $32$, $64$ and $128$. 
    \item \textbf{M1}: which uses $N_{re}=5$ with three stages of dimensions $64$, $128$ and $256$.
    \item \textbf{M2}: which uses $N_{re}=5$ with four stages of dimensions $64$, $128$, $256$ and $512$.
    \item \textbf{L}: which uses $N_{re}=7$ with four stages of dimensions $64$, $128$, $256$ and $512$.
\end{itemize}

\hfill

We use the M1 config on the Oxford-IIIT Pet dataset and the M2 configuration on the Kvasir-SEG and Cityscapes datasets.
We optimize the neural networks with AdamW~\cite{loshchilovdecoupled} with a learning rate of $\eta = 10^{-3}$, a batch size of $8$, and a weight decay of $10^{-4}$, we use a cosine annealing learning rate scheduler with $\eta_{\min} = 10^{-6}$, our networks are trained for 100 epochs on the Oxford-IIIT Pet dataset and for 200 epochs on the Kvasir-SEG and Cityscapes dataset. 
We use data augmentations to crop, flip and rotate images during training. For the temperature parameter $\tau$ of the scaled cross-entropy during training, we use $\tau=20.0$ on the Cityscapes dataset, on the Kvasir-SEG and Oxford-IIIT Pet dataset we use $\tau = 5.0$.

%
%


\section{Additional Experiments on the Kvasir-SEG Dataset}
\label{app:with_labels}

\begin{figure}[ht]
    \centering
    \includegraphics[width=0.95\linewidth]{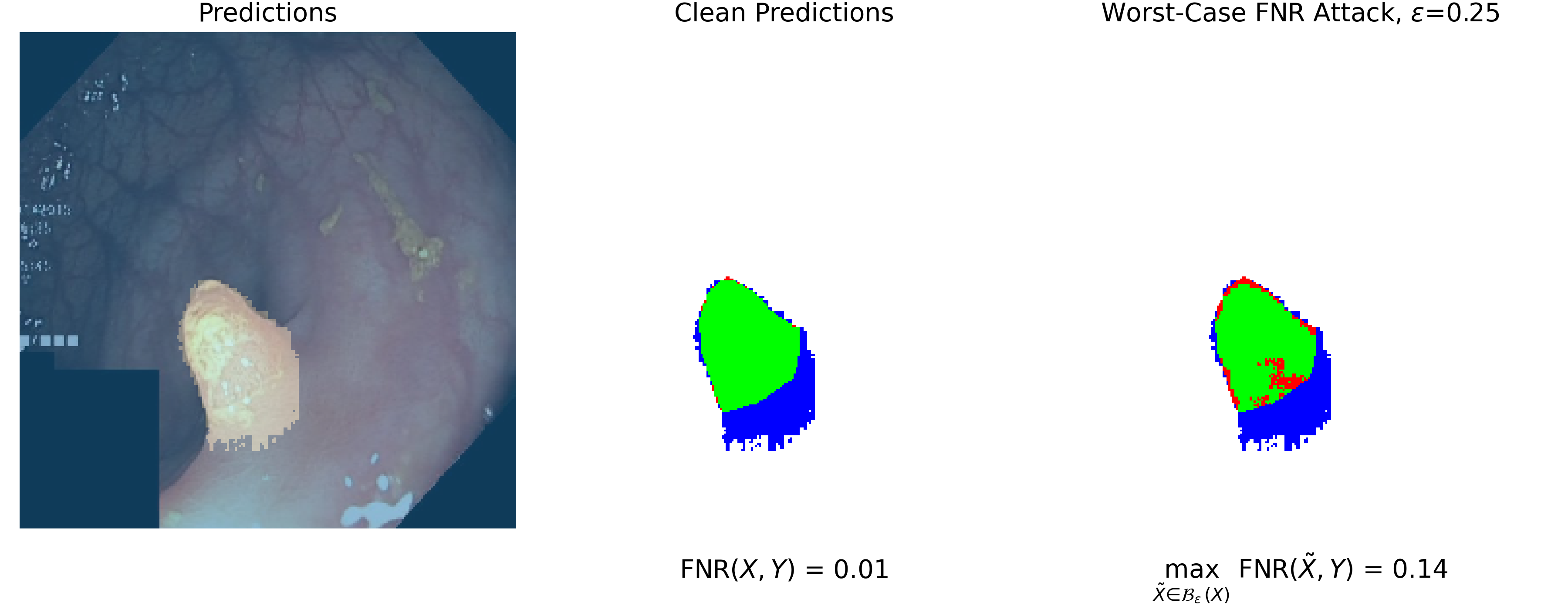}
    \caption{(left) The segmented output using our Lipschitz network, the ``polyp'' class is depicted in bright overlay. (middle) The predicted ``polyp'' pixels are shown as either \textcolor{green}{True Positives (TP) in green}, \textcolor{blue}{False Positives (FP) in blue}, or \textcolor{red}{False Negatives (FN) in red}. True Negatives (TN) are left in white. With very few FN, the FNR is very low (here, FNR = 0.01). (right) We depict the optimal output attack that maximizes the FNR for budget $\epsilon = 0.25$, which corresponds to the maximum number of attacked TP pixels. Therefore, we attack the less robust TP pixels in priority. We certify that we could not degrade the FNR more than 0.14. Note that this worst-case scenario is not necessarily feasible via an adversarial attack, however, it provides a useful tool for certification purposes.}
    \label{fig:with_labels}
\end{figure}

To further illustrate the different guarantees our framework allows on safety critical datasets, we provide segmentation results on the Kvasir-SEG dataset. Here, we use the S config of our DeepLabV3 architecture on the Kvasir-SEG dataset. For every image/label pair $X$/$Y$ in the dataset, we answer \textbf{Q1} by first computing the predictions via a simple forward pass of the network. Then, we compute the worst-case segmentation that would maximize the FNR measure under budget $\epsilon$. Results are presented in Fig.~\ref{fig:with_labels}.


\section{Guarantees in the Absence of Ground Truth Labels}

Previously, we showed how LipNets allowed for quick estimations of worst-case performance on semantic segmentation tasks. However, most of the guarantees provided by certifiable methods estimate the worst-case accuracy under attack. This requires access to ground-truth labels $Y$. We have to propose a way to convey a neural network's robustness at inference time in real-world deployment scenarios. In the main paper, we propose a stability performance measure that allows for the computation of robust pixel subsets, where more than $\alpha\%$ (or $N_\mathrm{max}$) predictions would not change under attacks with a budget inferior to a user-chosen $\epsilon$ in $\ell_2$ norm (see \cref{sec:other_metrics}). 

\begin{figure}[ht]
    \centering
    \includegraphics[width=0.65\linewidth]{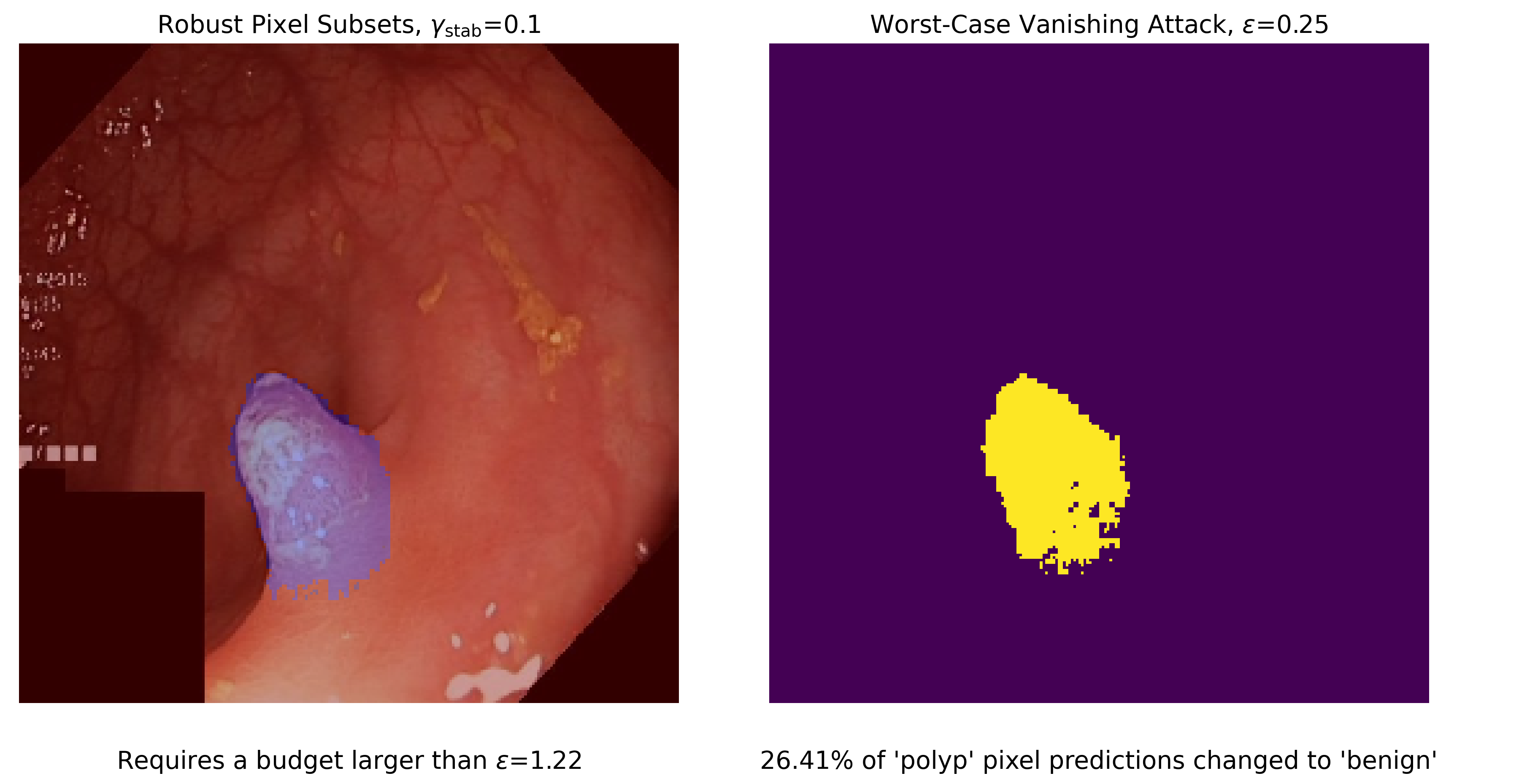}
    \caption{(left) We depict the subsets of adjacent pixel predictions of the class \textit{polyp} (one subset here). We also compute the necessary adversarial budget $\epsilon$ to perturb at least $90\%$ of pixel predictions from that subset of coordinates. (right) We display the worst-case vanishing attack under budget $\epsilon$, i.e, the perturbation that changes the maximum number of polyp predictions to benign predictions on the initial subset of identical predictions.}
    \label{fig:without_labels}
\end{figure}

Additionally, in terms of inference-time guarantees, we also provide a visual example of the stability of Lipschitz neural network predictions. In the same setting as Appendix~\ref{app:with_labels}, we provide some guarantees on the robustness of dense subsets of adjacent pixels that share the same classification (denoted $S_\omega$). We will measure their robustness to degradations with the following objective: 

\begin{equation}
\begin{aligned}
    & \kappa(\tilde{Y}, \hat{Y}) = \mathbbm{1}_{h(\tilde{Y}, \hat{Y}) \leq \gamma_\mathrm{stab}}. \\
    &\text{with} \ h(\tilde{Y}, \hat{Y}) = \sum_{\omega \in S_\omega} \mathbbm{1}_{\tilde{Y}_\omega = \hat{Y}_\omega}/|S_\omega|
\end{aligned}
\end{equation}

In Fig.~\ref{fig:without_labels}, we provide a visualization of inference results without assuming ground truth access. In this context, we provide a visualization of the network's predictions along with the worst-case outputs of the network under an attack of budget $\epsilon=0.25$ that aims to eliminate pixels of the class \textit{polyp}. We called it "vanishing attack" with $S_\omega = \{\omega \in \Omega, \hat{Y}_\omega = 1\}$.

\end{document}